\documentclass[11pt]{article}

\usepackage[final]{acl}

\usepackage{times}
\usepackage{latexsym}
\usepackage{amsmath} 
\usepackage{booktabs}
\usepackage{tabularx}
\usepackage[T1]{fontenc}
\usepackage{multirow}
\usepackage[utf8]{inputenc}
\usepackage{microtype}

\usepackage{inconsolata}

\usepackage{graphicx}

\title{ToolLoop: Closed-Loop Tool-Use Data Synthesis via Decomposed Generation and Dynamic Self-Feedback}

\author{
 \textbf{Min Zeng},
 \textbf{Yuzhou Liu},
 \textbf{Zhenyu Cao},
 \textbf{Hanxiu Chen},
 \textbf{Heng Li},
 \textbf{Caiquan Liu},
 \\
 \textbf{Yafei Wen},
 \textbf{Xiaoxin Chen}
\\
vivo AI Lab
\\
\small{
\href{mailto:zengmin325@163.com}{zengmin325@163.com}
}
  }

\begin{document}
\maketitle
\begin{abstract}
High-quality tool-use data is critical for training language models to interact effectively with external tools. However, existing synthetic approaches typically follow a generate-then-filter paradigm with static post-hoc verification, often yielding inefficient data with imbalanced feature distributions. We propose \textbf{ToolLoop}, a closed-loop framework that decomposes synthesis into three progressive stages: (1) sampling function name combinations as ground truth; (2) backward derivation of user queries; and (3) forward derivation of tool calls. At each stage, dynamic self-feedback iteratively guides the model toward high-quality generation, realizing a transition from generate-then-filter to generate-verify-refine. On the Berkeley Function Calling Leaderboard (BFCL), a 4B parameter model trained with our 11K synthetic examples achieves 86.40\% accuracy in non-reasoning mode, while an Isolate variant that removes BFCL-overlapping candidate functions still reaches 86.07\%. Cross-benchmark evaluation on ACEBench further demonstrates strong generalization, with 72.1\% overall accuracy using only 18.3\% of baseline training data.
\end{abstract}

\section{Introduction}
Large language models (LLMs) have demonstrated remarkable capabilities in natural language understanding and generation \citep{agashe2024agent, Ssearch,Optimized,Tapeagents}. However, they remain fundamentally constrained by their static knowledge and limited ability to interact with the external world \citep{c1,c2}. To address these limitations, recent research has adopted agent-based paradigms in which autonomous LLM-powered agents not only generate text but also interpret user intent and invoke external tools (e.g., APIs for weather queries, web search, or other services) to fulfill user requests, effectively bridging language understanding with real-world actions \citep{xu2025llm, wang2025enhancing,c3}. Despite this progress, training models to effectively leverage such tools requires high-quality tool-use datasets that accurately capture the relationship between user instruction and tool selection — a resource that remains extremely scarce in current datasets \citep{yang2025toolmind,c4}. 

Existing synthetic methods for tool-use data primarily follow a “generate-then-filter” paradigm, where complete samples are generated in a single step and subsequently filtered using rule-based or model-based verification \citep{xu2025toucan, liu2024toolace, ApiGen}. While straightforward, this approach suffers from three fundamental limitations. First, due to the diversity and complexity of tools, selecting the correct and most efficient tool from a large candidate set in a generation is highly challenging \citep{kim2025evaluating}. Second, static post-hoc filtering operates as a binary accept/reject mechanism that discards invalid instances without providing corrective feedback, thereby leading to biased feature distributions in the resulting dataset. Third, the absence of intermediate supervision during generation implies that the model receives no guidance on logical consistency between stages of generation (e.g., whether the generated tool call actually resolves the user query), leaving certain error modes unmitigated \citep{wang2025toolflow}. 

\begin{figure*}[htbp]
	\centerline{\includegraphics[width=0.9\textwidth]{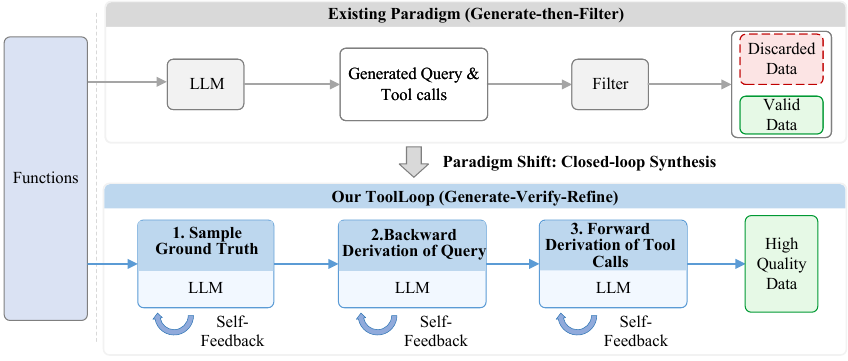}}
	\caption{Comparison between existing generate-then-filter paradigm and our ToolLoop closed-loop synthesis framework with generate-verify-refine approach.}
	\label{intro}
\end{figure*}

To address these challenges, as illustrated in Figure~\ref{intro}, we propose \textbf{ToolLoop}, a closed-loop data synthesis framework that rethinks the generation process through decomposed generation and dynamic self-feedback. ToolLoop decomposes synthesis into three progressive stages with explicit intermediate representations. At each stage, we integrate a dynamic self-verification mechanism that detects problematic outputs and feeds identified issues back to the language model as explicit guidance for regeneration. This closed-loop feedback enables the model to learn from its mistakes during synthesis and improves both efficiency and quality of retained samples. We conduct comprehensive evaluations on BFCL \cite{patil2025bfcl} and ACEBench \cite{chen2025acebench}, demonstrating strong improvements over existing approaches on both benchmarks.

Our contributions can be summarized as follows:
\begin{itemize}
    \item We introduce ToolLoop, a novel closed-loop framework that decomposes synthesis into three stages and integrates dynamic self-feedback to guide iterative optimization.
    \item We demonstrate through comprehensive experiments on BFCL and ACEBench that ToolLoop achieves strong downstream task performance with substantially less training data.
    \item We demonstrate the practical utility of ToolLoop by constructing an 11K-example synthetic tool-use dataset covering four representative function-calling scenarios.
\end{itemize}

\section{Related Work}

\noindent\textbf{Synthetic Data for LLMs.} Recent advances in tool-augmented language models have focused on improving both model capabilities and training data quality. Early synthetic-data methods primarily focused on general instruction tuning. Self-Instruct \citep{wang2023self} enabled language models to generate instruction--input--output examples, followed by filtering and deduplication, while Stanford Alpaca \citep{alpaca} built on this paradigm to improve synthesis diversity and efficiency. Subsequent work extended synthetic data generation to more diverse tasks and data types. Persona Hub \citep{ge2024scaling} introduced a persona-driven methodology for synthesizing reasoning data; APIGen \citep{ApiGen} automated the generation of verifiable function-calling data; SynthAgent \citep{wang2025adapting} improved synthetic data quality through the joint refinement of tasks and trajectories derived from web elements; and APIGen-MT \citep{prabhakar2025apigen} proposed a two-phase framework for generating verifiable and diverse multi-turn agent data.

\begin{figure*}[htbp]
	\centerline{\includegraphics[width=0.85\textwidth]{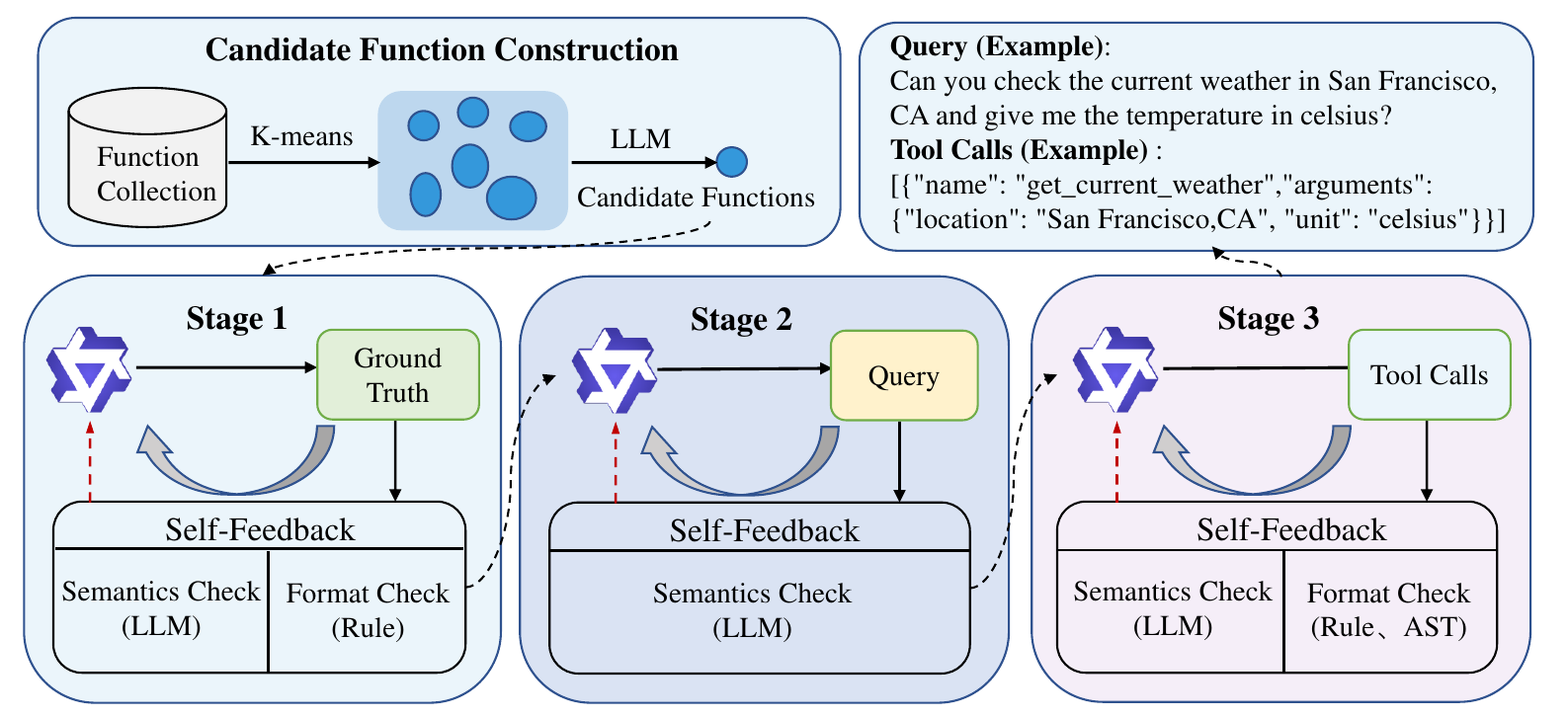}}
	\caption{ ToolLoop Framework: Three-Stage Decomposed Generation with Dynamic Self-Feedback}
	\label{ToolLoop}
\end{figure*}

\noindent\textbf{LLMs as Automatic Evaluators / Judges.}
LLMs are increasingly used as automatic evaluators (LLM-as-a-Judge) of model outputs. Laskar et al.\ \citep{Laskar2025LLMJudge} explore using LLMs as judges to evaluate biomedical relation extraction systems, demonstrating that structured output formatting can improve LLM-judge performance relative to traditional metrics. Similarly, Bavaresco et al.\ \citep{Bavaresco2025LLMvsHuman} provide a comprehensive empirical study across 20 NLP tasks, showing that LLMs can replicate some aspects of human annotation but also reveal substantial variability depending on task and model choice. D’Souza et al.\ \citep{DSouza2025YESciEval} introduce YESciEval, a robust LLM-as-a-Judge framework for scientific question answering with rubric-based assessment, supporting scalable evaluation without human feedback. Beyond systematic benchmarks, Yamauchi et al.\ \citep{Yamauchi2025EmpiricalJudge} empirically analyze how design choices in LLM-based evaluation affect alignment with human judgments, reporting important factors such as decoding and criteria formulation. Finally, Liu et al.\ \citep{Liu2025ConsJudge} propose Judge-Consistency for improving consistency of LLM judgments on retrieval-augmented generation. These works suggest that LLM-based assessment can serve as an effective, scalable alternative to traditional evaluation protocols in many scenarios.

\section{Methods}

\subsection{Scope of Category}
To ensure diversity and comprehensiveness in our synthesized data, we define a set of tool-call\footnote{In this paper, the terms tool, function, and API are used interchangeably unless otherwise specified.}
 scenarios that span different levels of complexity and patterns of use. These include: Simple, where a single provided function is invoked to answer the query; Multiple, where several functions are available but only one is needed; Parallel, where a single function must be called multiple times within one response and each call is independent; and Parallel Multiple, where the model selects one or more relevant functions from multiple candidates and may invoke them multiple times in parallel. Together, these scenarios cover the majority of single-turn tool-call cases, from straightforward single-function calls to complex multi-function parallel invocations.

\subsection{Candidate Function Construction}
Synthetic data requires constructing candidate functions for different tasks under each scenario. While random sampling and direct LLM-based filtering are potential approaches, they either fail to capture semantic relationships between tools or are limited by context window constraints when dealing with large function libraries. Based on this, we employ a method that combines semantic clustering with LLM-guided sampling. As shown in Figure~\ref{ToolLoop}, we first use a vector model to convert the textual description of each function into dense semantic embeddings. We then apply K-means clustering to partition the embedding space into K clusters, where each cluster contains semantically related functions that typically share the same domain or usage scenario. Finally, we sample functions belonging to the same usage scenario from each cluster via LLM as candidate functions for each Function invocation.

\subsection{Decomposed Data Synthesis with Dynamic Self-Feedback}
Given the substantial difficulty of directly synthesizing valid tool-use data in a single pass, we decompose the synthesis pipeline into three progressive stages: ground truth generation, user query derivation, and tool call instantiation. Each stage incorporates dynamic self-feedback to ensure quality before proceeding. Detailed prompt templates for all three stages with integrated self-feedback mechanisms are provided in Appendix~\ref{appb}. We describe each stage in detail below.

\subsubsection{Ground Truth Sampling}
A ground truth represents the sequence of function names that will be invoked to address a user's request. For parallel scenarios, we prompt an LLM to select function names from the candidate set that can potentially produce parallel invocations within a coherent usage context. For non-parallel scenarios, we employ random sampling to select function as the target invocation sequence.

We enforce the following properties during ground truth generation:

\begin{itemize}
    \item \textbf{Parallelizability}: Functions in the ground truth can be invoked concurrently without dependencies---no function requires the execution results of another.
    
    \item \textbf{Coherence}: The combination of functions should enable a reasonable user request within a unified scenario context.
    
    \item \textbf{Cardinality}: For parallel and parallel-multiple scenarios, the ground truth contains 2--4 function invocations, where the same function name may appear multiple times to support parallel calls. For simple and multiple scenarios, the ground truth contains a single target invocation.
\end{itemize}

\subsubsection{Backward Derivation of User Query}

Given a generated ground truth and the associated candidate functions, we perform backward reasoning to derive a natural user query. This reverse generation ensures tight alignment between user intent and the ground truth. We guide query generation with the following desiderata:

\begin{itemize}
    \item \textbf{Precision}: The implicit tool calls requirements in the query must correspond exactly to the ground truth---neither more nor fewer function calls should be implied.
    
    \item \textbf{Consistency}: Parameters mentioned in the query must conform to the function signatures and maintain format consistency across the query.
    
    \item \textbf{Completeness}: The query must provide all necessary information to satisfy the parameter requirements of the functions in the ground truth.
    
    \item \textbf{Naturalness}: The query should emulate realistic, conversational user expressions rather than mechanistic command-like instructions.
\end{itemize}

\subsubsection{Forward Derivation of Tool Calls}

In the final stage, we generate the concrete tool calls with fully specified parameters. Taking the user query and candidate functions as input, we prompt an LLM to produce the complete tool calls sequence in a structured format. We enforce the following constraints:

\begin{itemize}
    \item \textbf{Schema Compliance}: Generated tool calls must strictly adhere to the function signatures, with correct parameter names, types, and structures.
    
    \item \textbf{Functional Accuracy}: Tool calls must accurately fulfill the user's request as expressed in the query, with appropriate parameter values.
    
    \item \textbf{Format Specification}: We adopt the OpenAI function-calling format for standardized tool call representation, ensuring compatibility with existing evaluation frameworks.
\end{itemize}

\subsubsection{Dynamic Self-Feedback}

Critically, we integrate a dynamic self-feedback mechanism at each of the three stages described above. This mechanism combines three validation strategies: an LLM verifier assesses semantic correctness and logical coherence; deterministic rules check format compliance, structural constraints, and data types; and AST parsing detects syntactic errors in generated tool calls that would cause execution failures at runtime.

The feedback is stage-specific rather than a generic quality score. In ground truth sampling, the verifier is designed to assess whether the selected function names form a coherent intent, especially for parallel cases where calls should be independent but semantically related. In backward query derivation, it checks whether the user request is aligned with the intended function sequence and provides sufficient information for required arguments. In forward derivation, the checks become more concrete: the generated tool calls should follow the target schema and instantiate arguments using values supported by the user query. This stage-local design helps prevent early mistakes from being hidden until final filtering, where they are harder to diagnose.

When validation fails at any stage, we construct a refinement prompt that includes: (1) the original generation prompt, (2) the failed output as a negative example, and (3) specific issues identified by the validators with actionable feedback. For instance, if AST parsing detects a syntax error in a tool call, the feedback explicitly states "Syntax error: unmatched closing bracket in arguments field" rather than simply marking the sample as invalid. This feedback-augmented prompt guides the model to correct its mistakes in subsequent generation attempts.

This design also preserves useful diversity. A pure filtering pipeline tends to discard difficult samples, which can bias the retained data toward short queries, simple schemas, or overly explicit user requests. After a valid function combination has been selected, later-stage refinements attempt to repair the query or tool calls around that target rather than immediately discarding the sample. Only samples that remain invalid after the maximum number of refinement attempts are discarded. In this way, the feedback loop acts as a targeted correction mechanism rather than a coarse accept/reject gate.

The refinement process continues iteratively until either: (a) all validation checks pass, allowing progression to the next stage, or (b) a maximum of three retries is reached, at which point the sample is discarded. Thus, a sample may receive one initial generation and up to three feedback-guided regeneration attempts. This design balances quality improvement with computational efficiency—empirically, we find that most correctable errors are resolved within one or two retries, while samples requiring more retries often have fundamental semantic issues that are unlikely to be resolved through additional refinement.

\section{Experiments}

\subsection{Experimental Setup}

\noindent\textbf{Models.} To evaluate the quality of the synthetic data generated by our approach, we adopt Qwen3-4B-Instruct-2507 \footnote{\url{https://huggingface.co/Qwen/Qwen3-4B-Instruct-2507}} \citep{qwen3technicalreport} as the base model. This non-deliberative variant exhibits strong general-purpose capabilities and serves as a stable foundation for our experiments. We benchmark our method against a wide range of state-of-the-art models, including open-source variants such as Qwen3 series, LLaMA 4~\citep{meta2025llama}, Gemma3~\citep{team2025gemma}, and GLM 4.6~\citep{zhipu2025glm}, as well as commercial models such as GPT-5.2, Gemini-3-Pro, Claude-Opus-v4.5, Grok 4.1, and Amazon-Nova-2.

\begin{figure*}[t]
	\centerline{\includegraphics[width=0.7\textwidth]{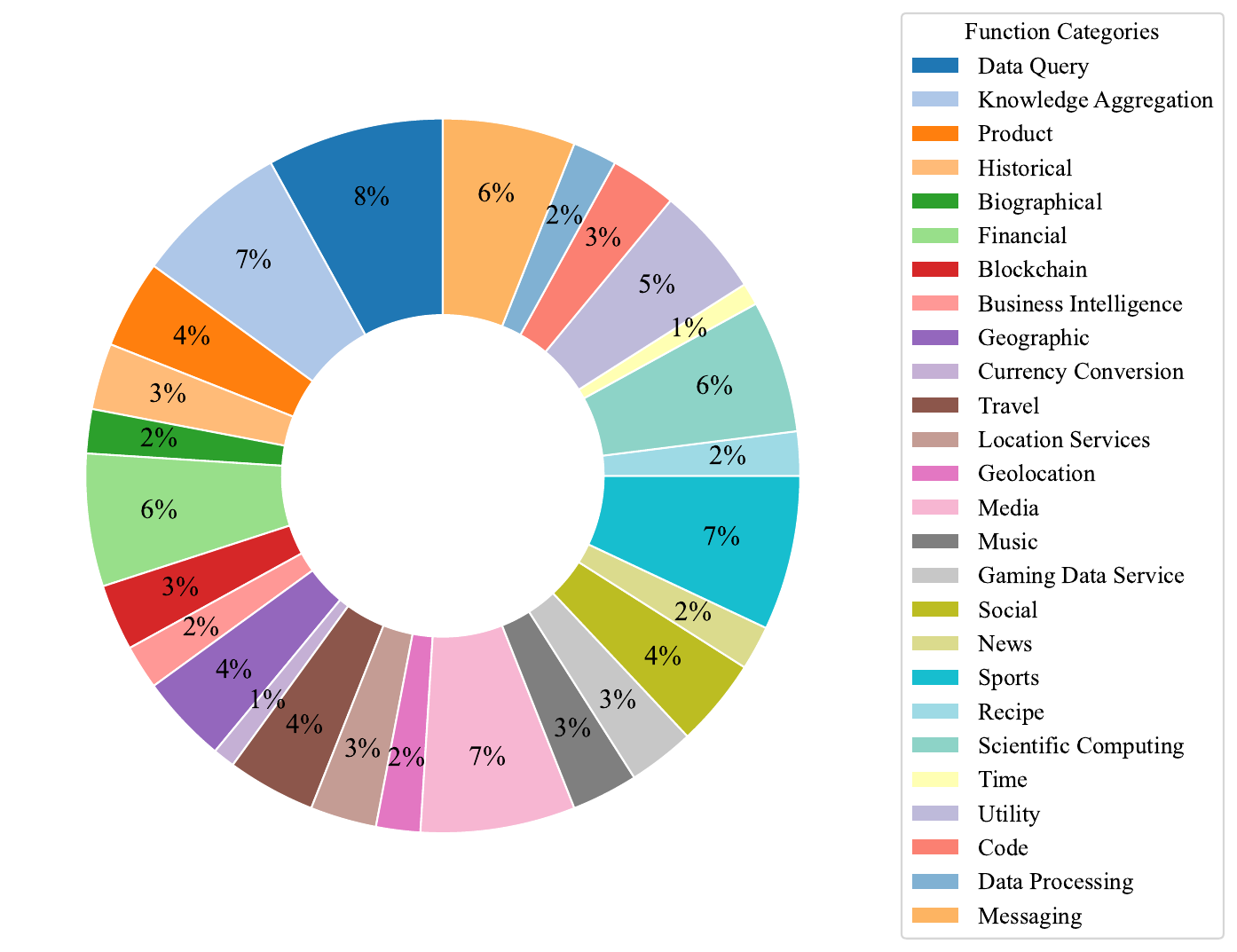}}
	\caption{Distribution of the executable APIs.}
	\label{apis}
\end{figure*}

\noindent\textbf{API Source.} Following the methodologies outlined in APIGen~\citep{ApiGen} and Magnet~\citep{magnet}, we randomly sample a subset of functions from ToolBench~\citep{qin2023toolllm} and BFCL to synthesize data. We collected a total of 5,281 executable APIs. For the leakage-control setting, ToolLoop-4B-Isolate filters candidate functions that overlap with BFCL evaluation candidate functions before synthesis, producing a 10K training set. These APIs were partitioned into 26 distinct semantic groups using K-means clustering applied to API descriptions encoded by the Qwen-3-Embedding-8B~\footnote{\url{https://huggingface.co/Qwen/Qwen3-Embedding-8B}} ~\citep{qwen3embedding}. We set the number of clusters to $K=26$ to ensure that each cluster contains approximately 200 data points; this size strikes a balance between effectively covering a rich diversity of data and staying within the text length constraints required for model inference. Figure~\ref{apis} demonstrates the distribution of the candidate API pool across diverse domains.

\noindent\textbf{Benchmark.} We evaluate the trained models on two complementary benchmarks:

\textit{BFCL-v4.} The Berkeley Function Calling Leaderboard~\citep{patil2025bfcl} contains 2,501 test instances across four scenarios: \textbf{Simple} (single function), \textbf{Multiple} (selecting from candidates), and \textbf{Parallel} (repeated invocations), \textbf{Parallel Multiple} (combined).

\textit{ACEBench.} ACEBench~\citep{chen2025acebench} is a comprehensive benchmark evaluating tool-use across five dimensions: \textbf{Atom} (atomic operations), \textbf{Single Turn} (complete task execution), \textbf{Similar API} (distinguishing similar tools), \textbf{Profile} (personalized selection), and \textbf{Overall} (aggregate performance). This provides a holistic view of model capabilities beyond function-calling accuracy.

\noindent\textbf{Hyperparameters.} All experiments are conducted on a multi-node cluster using the swift framework~\citep{zhao2025swift}, where each node is equipped with four NVIDIA L40s GPUs (48 GB VRAM each). The maximum sequence length for training data is set to 16k tokens. To mitigate overfitting, we train the models on synthetic datasets for two epochs.

For our ToolLoop variants, training uses the same base model, formatting convention, and non-reasoning inference setting, so the comparison between ToolLoop-4B and ToolLoop-4B-Isolate mainly reflects the effect of filtering BFCL-overlapping candidate functions rather than changes in decoding or model capacity. During evaluation, the model is required to output tool calls directly in the benchmark-compatible function-calling format, without additional chain-of-thought traces. This setting is intentionally strict: it measures whether the synthesized data teaches the model to map user intent to valid tool invocations, rather than whether the model can recover through verbose reasoning at inference time.

\begin{table*}[t]
    \centering
    \resizebox{1\textwidth}{!}{%
    \begin{tabular}{lccccccccccc}
        \toprule
        \multirow{3}{*}{Models} 
        & \multicolumn{4}{c}{Non-Live} 
        & \multicolumn{4}{c}{Live} 
        & \multicolumn{3}{c}{Overall} 
        \\
        \cmidrule(lr){2-5} \cmidrule(lr){6-9} \cmidrule(lr){10-12}
        & \multirow{2}{*}{Simple}  & \multirow{2}{*}{Multiple}  & \multirow{2}{*}{Parallel} & Parallel & \multirow{2}{*}{Simple} & \multirow{2}{*}{Multiple}  & \multirow{2}{*}{Parallel} & Parallel  & \multirow{2}{*}{Non-Live}  &  \multirow{2}{*}{Live}& \multirow{2}{*}{Overall}\\
         &  &  &  & Multiple & &  &  &  Multiple & &  & \\
        \midrule
        GPT-5.2-2025-12-11   & 72.92& 88.00 &89.00 &77.50 & 71.71 &70.37 &68.75 &58.33 & 81.85 & 70.39 & 76.12 \\
       Gemini-3-Pro-Preview & 75.50 & 94.00 & 91.00 & 82.50 & 87.60 & 80.44 & 75.00 & 79.17& 85.75& 81.72&83.74 \\
        Grok-4-1-fast & 77.58 & 93.00 & 92.50 & 90.00 & 84.11 & 77.30 & 75.00 & 70.83 &88.27& 78.46&83.37\\ 
        Claude-Opus-4-5-20251101 & 76.83 & 95.50 & 93.50 & 88.50 & 86.43 & 78.16 & 87.50 & 75.00 &88.58 & 79.79& 84.19\\
        Amazon-Nova-2-Lite-v1:0&76.33&94.00&91.50&86.00&83.33&80.15&87.50&79.17&86.96&80.83&83.90\\
        \midrule
        Qwen3-32B   & 75.58 & 94.50 & 93.50 & 91.50& \textbf{89.53} & \textbf{80.91} & 81.25 & 50.00 & 88.77& \textbf{82.01}&85.39 \\  
        Qwen3-4B-Instruct-2507   & 75.50 & 93.50 & 92.50 & 90.00 & 79.07 & 76.16 & 62.50 & 66.67&87.88&76.39&82.14 \\
        Llama-4-Scout-17B-16E  & 79.00 & 94.00 & 94.00 & 90.50 & 81.78 & 72.74 & 81.25 & 79.17&89.38&74.69&82.04 \\
        Gemma-3-27b-it & 77.67 & 92.50 & 89.00 & 89.50 & 84.50 & 72.46 & \textbf{93.75} & 45.83&87.17&74.54&80.86 \\
        GLM-4.6 &74.25&	95.00&	91.50&89.50&\textbf{89.53}&	78.92&	81.25&	75.00& 87.56& 80.90&84.23\\
        \midrule
        APIGen-4B (60K)    & 79.58 & 94.50 & \textbf{95.00} & 90.50 & 74.81 & 76.83 & 56.25 & \textbf{83.33} &89.90
 &76.31 &83.11 \\ 
        ToolMind-4B (55K)    & 77.92 & 96.50 & 93.50 & 90.00 & 84.50 & 76.35 & 62.50 & 66.67 &89.48
 &77.57 &83.53 \\ 
        ToolLoop-4B-Isolate (10K)   & 79.33 & \textbf{97.00} & 94.50 & 93.50 & 86.05 & 79.87 & 81.25 & 79.17&91.08&81.05&86.07 \\
        ToolLoop-4B (11K)   & \textbf{79.67} & 96.50 & 94.50 & \textbf{94.50} & 86.82 & 80.34 & 75.00 & 79.17&\textbf{91.29}&81.50&\textbf{86.40} \\
        \bottomrule
    \end{tabular}%
    }
    \caption{Comparison on the BFCL (last updated on 2025-12-16) across non-live and live scenarios. The best result within each category is highlighted in \textbf{bold}.}
    \label{bfcl}
\end{table*}

\subsection{Main Results}

Considering that agentic systems require high time efficiency during operation, we train and evaluate our model in non-reasoning mode without additional chain-of-thought overhead. Table~\ref{bfcl} compares ToolLoop with commercial, open-source, and data-centric baselines on BFCL. ToolLoop-4B, trained on only 11K synthetic examples (Appendix~\ref{appa}), achieves the best overall accuracy of 86.40\%, improving over APIGen-4B (83.11\%, 60K examples) by 3.29 points and ToolMind-4B (83.53\%, 55K examples) by 2.87 points. The leakage-control variant, ToolLoop-4B-Isolate, removes BFCL-overlapping candidate functions and still reaches 86.07\%, only 0.33 points below the full model, suggesting that the gains are not driven by candidate-function overlap.

In non-live scenarios, ToolLoop achieves 91.29\%, surpassing the strongest non-ToolLoop baseline, APIGen-4B (89.90\%), by 1.39 points. The advantage is visible in complex selection and parallelization settings: ToolLoop obtains 96.50\% on \textit{Multiple} and 94.50\% on \textit{Parallel\_Multiple}, while the Isolate variant even reaches 97.00\% on \textit{Multiple}. These results support the value of decomposing synthesis around explicit function combinations rather than relying on one-shot sample generation.


In live scenarios, ToolLoop remains competitive with the strongest commercial and open-source systems, reaching 81.50\% overall and clearly exceeding the data-centric baselines APIGen-4B (76.31\%) and ToolMind-4B (77.57\%). In \textit{Live-Parallel\_Multiple}, ToolLoop scores 79.17\% versus APIGen-4B's 83.33\%, a gap that amounts to a single sample difference on this 24-instance split. Given the small split size, this result has high sampling uncertainty and does not support a strong category-level conclusion.

The small gap between ToolLoop-4B and ToolLoop-4B-Isolate further indicates that ToolLoop does not simply benefit from memorizing benchmark-specific function schemas. After removing BFCL-overlapping candidate functions, the model retains nearly the same overall performance and even improves on the non-live \textit{Multiple} category. This suggests that the main benefit comes from the synthesis procedure itself: clustering related tools, constructing explicit target function combinations, and using feedback-guided regeneration to align user queries with executable tool calls. In other words, ToolLoop improves the model's general tool-selection and argument-grounding behavior rather than only adapting it to a particular function pool.

\subsection{Ablation Study}
To investigate the contribution of dynamic self-feedback, we compare three variants under identical training settings, as shown in Table~\ref{tab:ablation}. The \textit{w/ Final Filtering} variant uses the same synthesis framework, verifier, and initial generation budget as ToolLoop; its key difference is that invalid outputs are discarded rather than revised using verifier feedback, making it a controlled generate-then-filter baseline. Removing the feedback mechanism entirely (\textit{w/o Feedback}) yields only 79.97\% overall accuracy—a 2.17-point drop from the base model (82.14\%), suggesting that naively generated synthetic data without quality control introduces noisy training signals that actively harm model performance. Replacing iterative refinement with static post-hoc filtering partially recovers performance (82.56\%), yet still falls far short of the full system, as discarding invalid samples without corrective guidance leaves systematic generation errors unaddressed. In contrast, the full ToolLoop framework achieves 86.40\% overall (91.29\% non-live, 81.50\% live), demonstrating that the key to high-quality synthetic data lies not in generation alone, but in the iterative \emph{generate-verify-refine} loop enabled by dynamic self-feedback.

\begin{table}[t]
\centering
\small
\resizebox{\columnwidth}{!}{%
\begin{tabular}{lccc}
\toprule
\textbf{Model Variant} & \textbf{Non-Live} & \textbf{Live} & \textbf{Overall} \\
\midrule
Qwen3-4B-Instruct (Base) & 87.88 & 76.39 & 82.14 \\
\quad w/o Feedback & 85.02 & 74.97 & 79.97 \\
\quad w/ Final Filtering & 88.81 & 76.31 & 82.56 \\
\quad w/ Feedback (ToolLoop) & \textbf{91.29} & \textbf{81.50} & \textbf{86.40} \\
\bottomrule
\end{tabular}
}
\caption{Ablation study of ToolLoop components on BFCL.}
\label{tab:ablation}
\end{table}

This ablation highlights a practical distinction between filtering and refinement. Final filtering can remove malformed samples, but it cannot recover cases where the user query, target function sequence, and final arguments are individually plausible yet mutually inconsistent. By providing feedback at each intermediate stage, ToolLoop corrects these alignment errors before they propagate to later steps, which explains why the full framework improves both non-live and live performance rather than merely increasing the number of accepted samples.

\subsection{Generalization to Broader Tool-Use Benchmarks}

\begin{table*}[t]
\centering
\small
\begin{tabular}{lccccc}
\toprule
\textbf{Model} & \textbf{Overall} & \textbf{Atom} & \textbf{Single Turn} & \textbf{Similar API} & \textbf{Profile} \\
\midrule
Qwen3-4B-Instruct-2507 & 64.9 & 68.0 & 59.5 & 68.0 & \textbf{64.0} \\
APIGen-4B (60K) & 67.0 & 76.0 & 62.0 & 76.0 & 54.0 \\
ToolMind-4B (55K) & 70.2 & 83.3 & \textbf{69.5} & 74.0 & 54.0 \\
ToolLoop-4B (11K) & \textbf{72.1} & \textbf{84.0} & 66.5 & \textbf{78.0} & 60.0 \\
\bottomrule
\end{tabular}
\caption{Performance comparison on ACEBench across five evaluation categories.}
\label{acebench}
\end{table*}

To assess cross-benchmark generalization, we evaluate ToolLoop on ACEBench, which covers atomic operations, single-turn interactions, API similarity, and profile-based tool selection. To keep the comparison controlled, Table~\ref{acebench} focuses on the base model and two data-centric baselines, APIGen-4B and ToolMind-4B.

ToolLoop-4B achieves the best overall score of 72.1\%, improving over APIGen-4B by 5.1 points while using only 18.3\% of its training data, and over ToolMind-4B by 1.9 points while using one fifth of its data. It also performs best on \textit{Atom} (84.0\%) and \textit{Similar API} (78.0\%), indicating that ToolLoop improves both basic tool-call semantics and fine-grained tool discrimination. These gains show that the closed-loop synthesis process transfers beyond BFCL-specific formatting.

The category-level results also reveal useful limits. In \textit{Single Turn}, ToolLoop reaches 66.5\%, outperforming APIGen-4B and the base model but trailing ToolMind-4B (69.5\%), making single-turn intent following a useful direction for improvement. In \textit{Profile}, all fine-tuned models underperform the base model, suggesting that synthetic tool-call fine-tuning can weaken personalized selection when user preferences are not explicitly modeled. ToolLoop nevertheless scores 60.0\%, higher than APIGen-4B and ToolMind-4B (both 54.0\%).

\section{Analysis}

\subsection{Synthesis Efficiency}
Table~\ref{tab:retries} presents the distribution of refinement iterations across the three progressive stages of ToolLoop synthesis. Stage 2 (Backward Query Derivation) has the highest refinement demand, with 18.1\% of samples requiring at least one retry. Compared with generate-then-filter pipelines, where failed samples are discarded and regenerated from scratch, ToolLoop concentrates its extra cost on the small fraction of samples requiring two or three retries: 1.72\% in Stage 1, 8.05\% in Stage 2, and 4.25\% in Stage 3. This suggests that closed-loop refinement adds modest overhead while substantially improving data quality.

Across the 11,024 retained examples, we estimate that synthesis consumed 26.13M input tokens and 7.28M output tokens, totaling 33.41M tokens as measured with the Qwen tokenizer. The cost varies by category because simple and multiple examples bypass LLM-based ground-truth sampling in Stage 1, and the Stage 2 query generator receives only the functions involved in the sampled tool chain rather than the full candidate pool. Parallel examples account for the largest share of the total cost because they require LLM-based tool-chain construction and typically contain longer function combinations. This accounting quantifies the computational trade-off of refinement: additional inference is concentrated on invalid intermediate outputs instead of regenerating every rejected example from scratch.

The higher retry rate in Stage 2 is expected because query generation is the point where symbolic function plans must be translated into natural language. A query can be fluent but still invalid if it omits a required parameter, implies an extra tool call, or describes a scenario that no longer matches the sampled function set. These errors are difficult for rule-based checks alone because they require semantic comparison between the query and the target function sequence. The relatively low retry rate in Stage 3 suggests that once the query is well aligned with the ground truth, producing schema-compliant tool calls becomes easier, especially with deterministic checks guarding argument names and structural validity.

This pattern also clarifies why ToolLoop gains more in complex scenarios. Multiple and parallel-multiple cases require the model to distinguish relevant tools from distractors while maintaining a consistent user intent across several calls. If the synthetic query is even slightly underspecified, the final trained model may learn ambiguous tool-selection behavior. By resolving such issues during data synthesis, ToolLoop increases the density of training examples that expose the model to hard selection decisions without introducing inconsistent supervision.
\begin{table}[h]
\centering
\small
\begin{tabular}{lcccc}
\toprule
Stage & \multicolumn{4}{c}{Number of Retries} \\
\cmidrule(lr){2-5}
 & 0 & 1 & 2 & 3 \\
\midrule
Stage 1 & 10720 & 114  & 161 & 29  \\
Stage 2 & 9029  & 1108 & 530 & 357 \\
Stage 3 & 10257 & 299  & 298 & 170 \\
\bottomrule
\end{tabular}
\caption{Distribution of refinement iterations per stage during ToolLoop 
         synthesis.}
\label{tab:retries}
\end{table}

We also examined the 280 samples that remained invalid after the maximum of three retries. Among them, 192 repeatedly failed LLM-based semantic verification, 83 failed deterministic rule-based validation, and 5 failed both types of checks. Most discarded cases therefore exhibited persistent semantic inconsistencies rather than only isolated formatting errors. This discarded set is small relative to the 11,024 retained examples, suggesting that the retry limit primarily removes unresolved supervision signals while iterative correction preserves most initially invalid samples.

\subsection{Verifier Reliability}

All three stages employ the same verifier model, Qwen-Max, as the LLM judge for semantic validation. To assess the reliability of this automated verifier, we manually annotated a random sample of 100 instances and compared the results against the model's judgments, yielding a human-agreement rate of 94\%. We treat this result as a sanity check for the semantic verifier rather than as a full validation of every error type or synthesis stage.

We use this manual annotation only as an overall reliability check rather than as a fine-grained error taxonomy. Among the few disagreement cases, errors were mainly associated with parameter-level constraints: for example, when a function schema required an integer argument, the verifier occasionally failed to flag cases where the generated tool call supplied a floating-point value with a fractional part. Such type-related errors can typically be captured by deterministic schema or AST-based checks. We also observed a small number of semantic grounding errors, such as treating a geographic location name as a city when the tool expected a city-level argument; unlike structural or type errors, these cases are harder to detect with AST parsing alone. These disagreements were infrequent and did not indicate a systematic failure pattern. The 94\% agreement rate supports the use of Qwen-Max as a semantic verifier, while rule-based and AST-based checks remain necessary for objective constraints such as JSON validity, argument names, data types, and schema conformance. Although this sample size does not provide a complete validation of all error categories, it serves as a sanity check that the verifier is sufficiently reliable for semantic screening.

\section{Conclusion}
This paper presents ToolLoop, a closed-loop framework for synthesizing high-quality tool-use data through decomposed generation and dynamic self-feedback. By decomposing synthesis into ground truth generation, user query derivation, and tool call instantiation, ToolLoop moves tool-use synthesis from generate-then-filter toward generate-verify-refine. Experiments on BFCL show that a 4B model trained on only 11K ToolLoop examples achieves 86.40\% accuracy in non-reasoning mode, while the BFCL-overlap-filtered Isolate variant still reaches 86.07\%. On ACEBench, ToolLoop achieves the best overall score among data-centric methods, suggesting that the synthesized data can transfer beyond the primary benchmark.

Taken together, our results suggest that the main value of ToolLoop lies in improving the internal consistency of synthetic examples rather than simply scaling the amount of supervision. By jointly verifying the intended function sequence, the derived user intent, and the final executable tool calls, ToolLoop produces training examples with better alignment across intermediate components. These findings support a simple takeaway: effective tool-use data should be constructed through an iterative generate--verify--refine process rather than generated as a single query--answer pair and filtered only at the end. ToolLoop provides one practical implementation of this process and offers a starting point for building more reliable tool-use training data.

\section*{Limitations}
The absence of real environment feedback means we cannot verify whether our synthesized data adequately prepares models for handling practical challenges such as timeout errors, malformed API responses, or cascading failures in ground truths. Furthermore, real-world tool-use often involves iterative refinement based on execution results, a capability not assessed in current static evaluation protocols. Future work should establish interactive testbeds with executable tool environments to validate the practical applicability of synthetic training data.

ToolLoop also uses Qwen-Max as the semantic verifier across all three synthesis stages. Although the verifier achieves 94\% agreement with human annotations on a random sample of 100 instances and is complemented by deterministic schema and AST checks, this evaluation does not rule out verifier-specific or correlated semantic biases. Future work should compare independent verifier models, report stage-level calibration, and incorporate executable environment feedback.

\section*{Ethics Statement}
This work focuses on synthetic data generation for tool-use in language models and does not collect personally identifiable information or conduct user studies. The manual verifier-reliability assessment annotates only synthetic examples. All experiments use publicly available API specifications and benchmarks.

\bibliography{custom}
\appendix
\section{The distribution of synthetic data}
\label{appa}

\begin{figure}[htbp]
	\centerline{\includegraphics[width=0.49\textwidth]{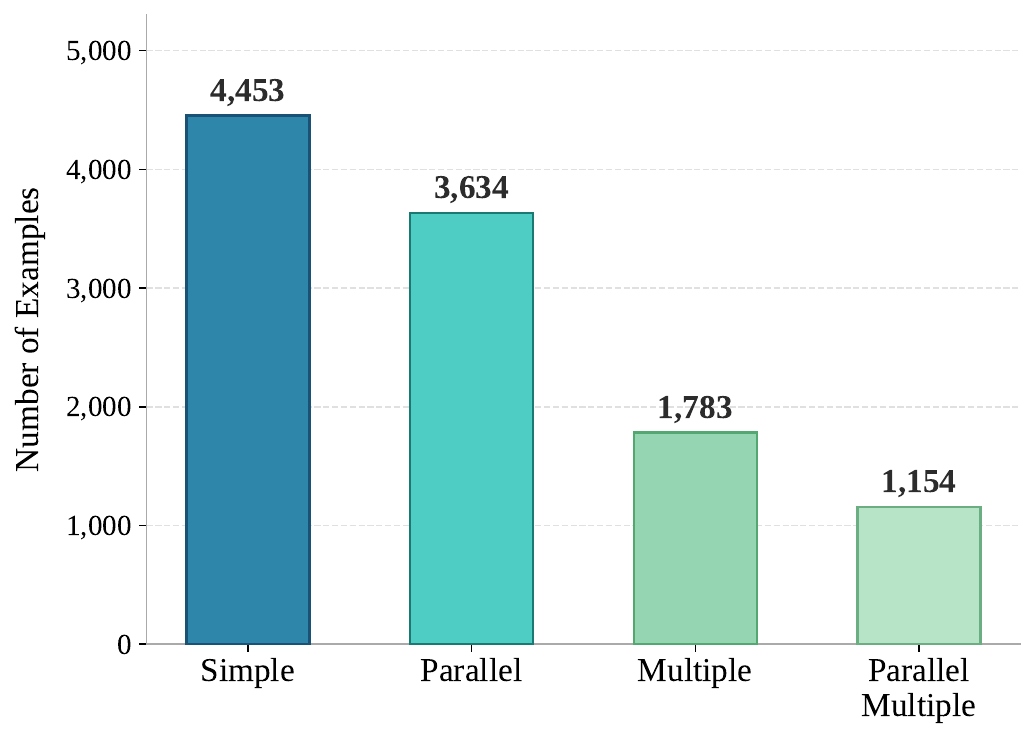}}
	\caption{Distribution of 11K synthetic training examples across four categories.}
	\label{data_distribution}
\end{figure}

Through the ToolLoop approach, we synthesized 11K data points, whose distribution across four query categories is shown in Figure~\ref{data_distribution}. Out of the 11,024 total examples, the dataset consists of 4,453 simple (40.4\%), 3,634 parallel (33.0\%), 1,783 multiple (16.2\%), and 1,154 parallel-multiple (10.5\%) instances.

\section{Prompt Design}
\label{appb}
\begin{figure*}[htbp]
	\centerline{\includegraphics[width=0.9\textwidth]{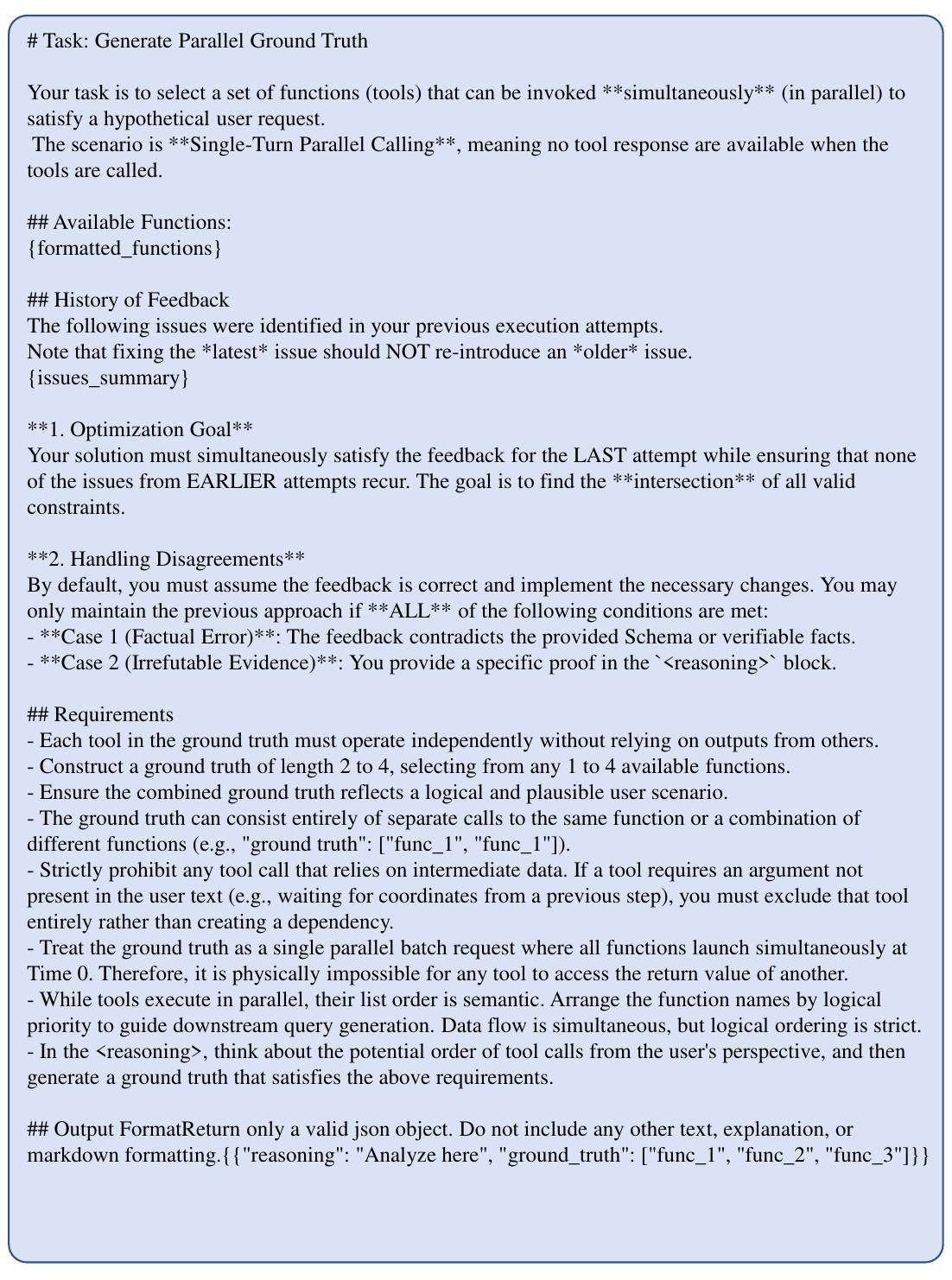}}
	\caption{Prompt template for Stage 1: Ground Truth Sampling with dynamic self-feedback.}
	\label{p1}
\end{figure*}

\begin{figure*}[htbp]
	\centerline{\includegraphics[width=0.9\textwidth]{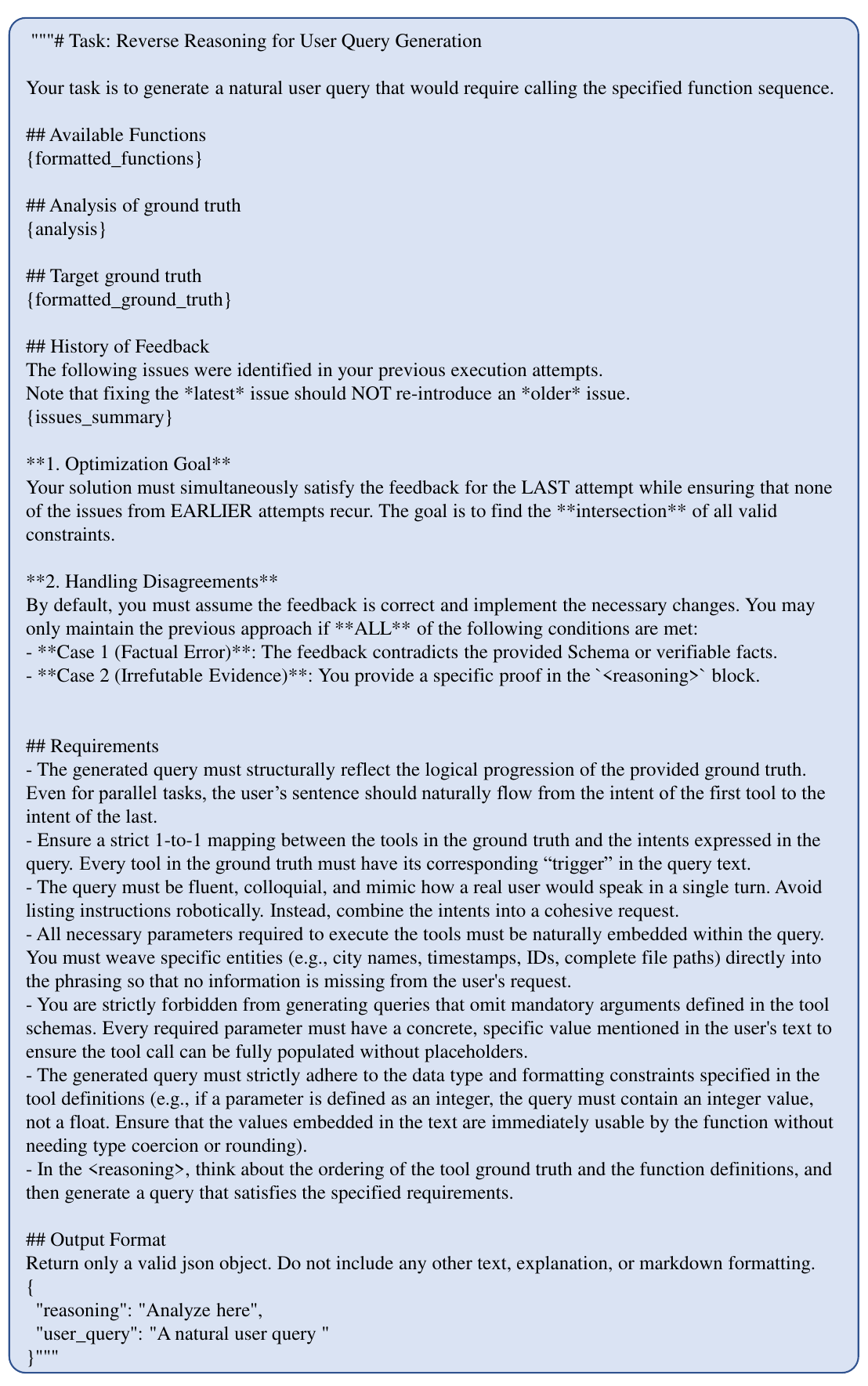}}
	\caption{Prompt template for Stage 2: Backward Derivation of User Query with dynamic self-feedback.}
	\label{p2}
\end{figure*}

\begin{figure*}[htbp]
	\centerline{\includegraphics[width=0.9\textwidth]{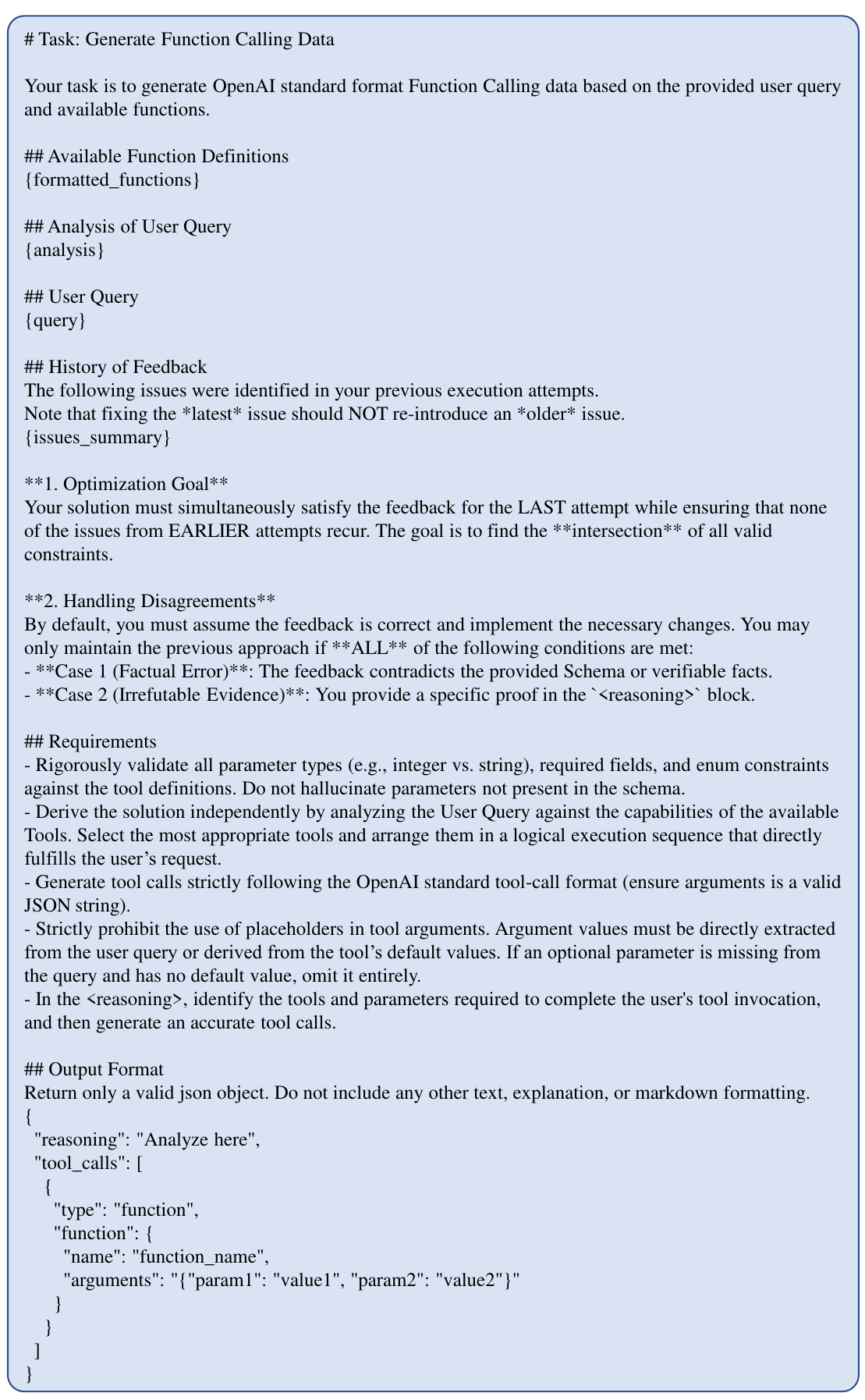}}
	\caption{Prompt template for Stage 3: Forward Derivation of Tool Calls with dynamic self-feedback.}
	\label{p3}
\end{figure*}

To ensure reproducibility and transparency of our approach, we provide the complete prompt templates used in each stage of ToolLoop's decomposed generation process. Figures~\ref{p1}, \ref{p2}, and \ref{p3} present the detailed prompts for Stage 1 (Ground Truth Sampling), Stage 2 (Backward Derivation of User Query), and Stage 3 (Forward Derivation of Tool Calls), respectively. Each prompt incorporates our dynamic self-feedback mechanism through the History of Feedback section, which guides iterative refinement based on identified issues from previous attempts.

\end{document}